\documentclass{article}

\usepackage{arxiv}

\usepackage[utf8]{inputenc} 
\usepackage[T1]{fontenc}    
\usepackage{hyperref}       
\usepackage{url}            
\usepackage{booktabs}       
\usepackage{amsfonts}       
\usepackage{nicefrac}       
\usepackage{microtype}      
\usepackage{lipsum}		
\usepackage{graphicx}
\usepackage{natbib}
\usepackage{doi}

\usepackage{amsmath}
\usepackage{booktabs}
\usepackage{xcolor}
\usepackage{soul}
\sethlcolor{pink}

\usepackage{subcaption}

\usepackage{listings}

\usepackage{xspace}
\newcommand{\gemma}{Gemma\xspace}
\newcommand{\llama}{Llama\xspace}
\newcommand{\mistral}{Mistral\xspace}
\newcommand{\qwenfamily}{Qwen3\xspace}

\newcommand{\qwenone}{Qwen3-1.7B\xspace}
\newcommand{\qwenfour}{Qwen3-4B\xspace}
\newcommand{\qweneight}{Qwen3-8B\xspace}
\newcommand{\qwenfourteen}{Qwen3-14B\xspace}

\newcommand{\gemmafull}{Gemma-2-9B-It\xspace}
\newcommand{\llamafull}{Llama-3-8B-Instruct\xspace}
\newcommand{\mistralfull}{Mistral-7B-Instruct-v0.3\xspace}

\title{When Text and Numbers Disagree: Evidence Arbitration in Large Language Models}

\author{
\textbf{Mattia Carletti}$^{1}$,
\textbf{Edward Phillips}$^{1}$,
\textbf{Fredrik K. Gustafsson}$^{1}$,
\textbf{Patitapaban Palo}$^{1}$,\\
\textbf{Lei Clifton}$^{2}$,
\textbf{Danielle Belgrave}$^{3}$,
\textbf{Xiao Gu}$^{1}$,
\textbf{David A. Clifton}$^{1,4}$\\[0.7em]
$^{1}$Department of Engineering Science, University of Oxford, Oxford, UK\\
$^{2}$Nuffield Department of Primary Care Health Sciences, University of Oxford, Oxford, UK\\
$^{3}$GlaxoSmithKline, London, UK\\
$^{4}$Oxford Suzhou Centre for Advanced Research, University of Oxford, Suzhou, Jiangsu, China
}

\date{}

\renewcommand{\shorttitle}{When Text and Numbers Disagree}

\hypersetup{
pdftitle={When Text and Numbers Disagree: Evidence Arbitration in Large Language Models},
pdfsubject={q-bio.NC, q-bio.QM},
pdfauthor={},
pdfkeywords={},
}

\begin{document}
\maketitle

\begin{abstract}
Large language models (LLMs) are increasingly used in settings where textual summaries, numerical observations, and external tool outputs may provide conflicting evidence. We study how LLMs arbitrate between such sources when they support opposing decisions. To do so, we introduce a controlled synthetic benchmark in which latent risk trajectories generate both numerical time series and natural language summaries, allowing us to construct conflicts where exactly one evidence source is aligned with the ground-truth label. This design lets us independently manipulate modality, temporal recency, source reliability, and evidence provenance. Across open-weight instruction-tuned models, we find that arbitration behaviour is systematic rather than random: models exhibit distinct text-versus-number preferences, follow temporal recency more consistently than explicit reliability cues, and can over-rely on external forecasts even when they conflict with direct contextual evidence. These results suggest that current LLMs often rely on heuristic arbitration strategies when integrating heterogeneous evidence, highlighting a failure mode for tool-augmented decision systems.
\end{abstract}


\section{Introduction} 
\begin{figure}[t]
  \centering
  \includegraphics[scale=1]{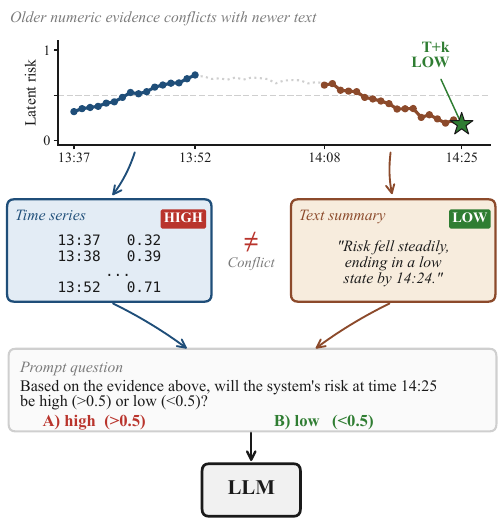}\vspace{1.0mm}
  \caption{%
    \textbf{Evidence arbitration under conflict.}
    A numerical time series and a textual summary, drawn from different windows of a shared latent risk trajectory, support \emph{conflicting} predictions at the target time $T{+}k$: $\textsc{high}$ for the older numerical source, $\textsc{low}$ for the newer textual source. The task requires the model to decide which source to prioritize when answering the binary-choice prompt.
  }
  \label{fig:pipeline}
\end{figure}

In many real-world decision-making scenarios, different sources of evidence may support conflicting conclusions. In healthcare, for example, a clinician’s assessment may describe a patient as stable while vital signs indicate ongoing deterioration. Similarly, in manufacturing, sensor readings may suggest normal machine operation even as maintenance reports point to impending failure. Such inconsistencies can arise for several reasons, including temporal mismatches between data sources (Figure~\ref{fig:pipeline}), differences in source reliability, or the integration of observational evidence with inaccurate predictions generated by external tools.

Large Language Models (LLMs) are increasingly being explored in high-stakes domains such as healthcare, mental health, and finance \citep{burton2024large, johri2025evaluation, goh2025gpt, heinz2025randomized, wang2026mental, xie2024finben, hu2026fin}, while also being incorporated into decision-making and agentic systems that integrate external tools and heterogeneous data sources \citep{qin2024toolllm, li-2025-review, yao2022react}. Understanding how these models behave when confronted with conflicting evidence is therefore increasingly important.

In this work, we use the term \emph{arbitration} to refer to how models prioritize or reconcile competing signals when different sources support incompatible conclusions. Failures of arbitration may lead models to privilege stale, unreliable, or incorrect tool-generated evidence over more relevant observations, producing unreliable decisions even when the correct signal is present in the prompt.

Prior work has studied conflicts between textual sources \citep{jiayang2024econ, kurfali-2025-conflicting}, between parametric and external knowledge \citep{xu2024knowledge, wang2023resolving, shi2024ircan, sui2025bridging, khandelwal-etal-2025-cocoa}, and across modalities such as image and text \citep{liu-etal-2025-insight, zhang2025modalities, jia2026benchmarking, shao2025cognition}. Other work has examined numerical reasoning \citep{li2025exposing, lovering2025language}, time-series forecasting \citep{gruver2023large, jin2024time, jia2024gpt4mts, liu2025calf}, and tool use in LLMs \citep{qin2024toolllm, qu2025tool, feng2025retool, li-2025-review}. However, these lines of work do not directly characterize how LLMs arbitrate between textual and numerical evidence, when these two support opposing decisions. This setting is increasingly relevant in applications where natural language summaries, numerical measurements, and external model or tool outputs are presented together.

In practice, conflicts between numerical and textual evidence are rarely attributable to modality alone. Instead, they arise from the interaction of multiple cues, including modality priors, temporal recency, source reliability, and evidence provenance (direct observations versus externally generated predictions). These cues can point in different directions: a newer textual report may contradict older numerical measurements, a reliable time series may conflict with a corrupted summary, or an external forecast may disagree with directly observed context. Systematically characterizing such behaviour is difficult in unconstrained real-world data because the reliability, provenance, and ground truth associated with each source are often ambiguous. To address this, we introduce a controlled synthetic benchmark in which these properties are known by construction. 

The benchmark uses latent risk trajectories to generate both numerical time series and natural language summaries. This framework allows us to construct conflicts where exactly one evidence source is aligned with the ground-truth label, while independently manipulating modality, temporal recency, source reliability, and evidence provenance. Our objective is not to reproduce the full complexity of deployed decision-making systems, but to isolate arbitration tendencies that may otherwise be difficult to identify in real-world environments.

Across open-weight instruction-tuned models, we find that arbitration behaviour is systematic rather than random. Models exhibit distinct text-versus-number preferences, follow temporal recency more consistently than explicit reliability cues, and can over-rely on external forecasts even when they conflict with direct contextual evidence. These findings suggest that current LLMs often rely on heuristic arbitration strategies when integrating heterogeneous evidence.

Our contributions can be summarized as follows:
\begin{itemize}
    \item We formulate \emph{textual--numerical evidence arbitration} as a controlled evaluation setting for studying how LLMs prioritize conflicting textual and numerical evidence.
    
    \item We introduce a large-scale synthetic benchmark that disentangles the effects of modality, temporal recency, source reliability, and evidence provenance on model decisions.

    \item We uncover systematic arbitration biases and failure modes across modern LLM families, including modality preferences, prompt-order effects, and over-reliance on external forecasts, with implications in settings involving heterogeneous or tool-derived evidence.
\end{itemize}

\section{Related Work}

\paragraph{Conflict Resolution and Evidence Arbitration.} A growing body of work has examined how LLMs handle conflicting information across different sources of knowledge. In retrieval-augmented and contextual generation settings, several studies investigate conflicts between parametric knowledge encoded in model weights and contextual knowledge provided at inference time and how LLMs behave under such discrepancies \citep{wang2023resolving, xu2024knowledge}. 

Building on this, recent approaches proposed mechanisms to improve conflict handling, including context-aware neuron reweighting \citep{shi2024ircan}, controlled integration of retrieved evidence through shared-private semantic modeling \citep{sui2025bridging}, and adaptive decoding strategies \citep{khandelwal-etal-2025-cocoa}. Related work on text-only evidence conflicts further shows that LLMs often exhibit strong positional and stylistic biases when resolving contradictions and rarely express uncertainty in the presence of conflicting evidence \citep{jiayang2024econ, kurfali-2025-conflicting}.

More recently, researchers have extended the study of knowledge conflicts to multimodal settings, particularly inconsistencies between visual evidence and internal commonsense or textual reasoning \citep{liu-etal-2025-insight, zhang2025modalities, shao2025cognition, jia2026benchmarking}. 

In the context of evidence arbitration in LLMs, our work addresses the critical yet largely unexplored challenge of textual--numerical conflicts.
\paragraph{LLMs for Numerical Tasks.}
Recent work has explored how LLMs can be adapted to numerical and forecasting tasks through reprogramming, multimodal prompting, and cross-modal alignment techniques \citep{gruver2023large,jin2024time,jia2024gpt4mts,langer2025opentslm,liu2025calf}. At the same time, several studies question the effectiveness of LLMs for numerical reasoning and forecasting, highlighting issues such as poor calibration, sensitivity to noise, weak temporal reasoning, and limited numerical understanding \citep{tan2024language,park2025revisiting,li2025exposing,lovering2025language}. Motivated by these limitations, we formulate our setting as a binary forecasting task rather than a purely numerical prediction problem, allowing us to study evidence arbitration without requiring precise numerical reasoning.
\paragraph{Tool-Augmented and Agentic LLM Systems.}
The rapidly growing subfield of tool-augmented and agentic LLMs has emphasized the role of iterative reasoning, planning, feedback, and external tool use in complex decision-making tasks \citep{yao2022react,qin2024toolllm,qu2025tool,feng2025retool,li-2025-review}. More recently, these paradigms have been extended to time series analysis, where LLM agents integrate textual reasoning with numerical and domain-specific evidence for forecasting and multi-step inference \citep{wang2024news,ye2026ts,zhao2025timeseriesscientist}. Our work is closely related to this line of research, where models often need to integrate numerical evidence with language-based reasoning.

\section{Textual--Numerical Evidence Arbitration}
We now define the arbitration task and the controlled benchmark used to evaluate it. Each instance asks a model to make a binary prediction about a future target value from textual, numerical, and optionally tool-derived evidence. In the conflict settings, two evidence sources support opposing labels, with exactly one source aligned with the ground truth. We then describe the four conflict dimensions, the synthetic data and prompt generation pipeline, and the evaluation protocol.

\begin{figure*}[t]
    \centering
    \includegraphics[width=\linewidth]{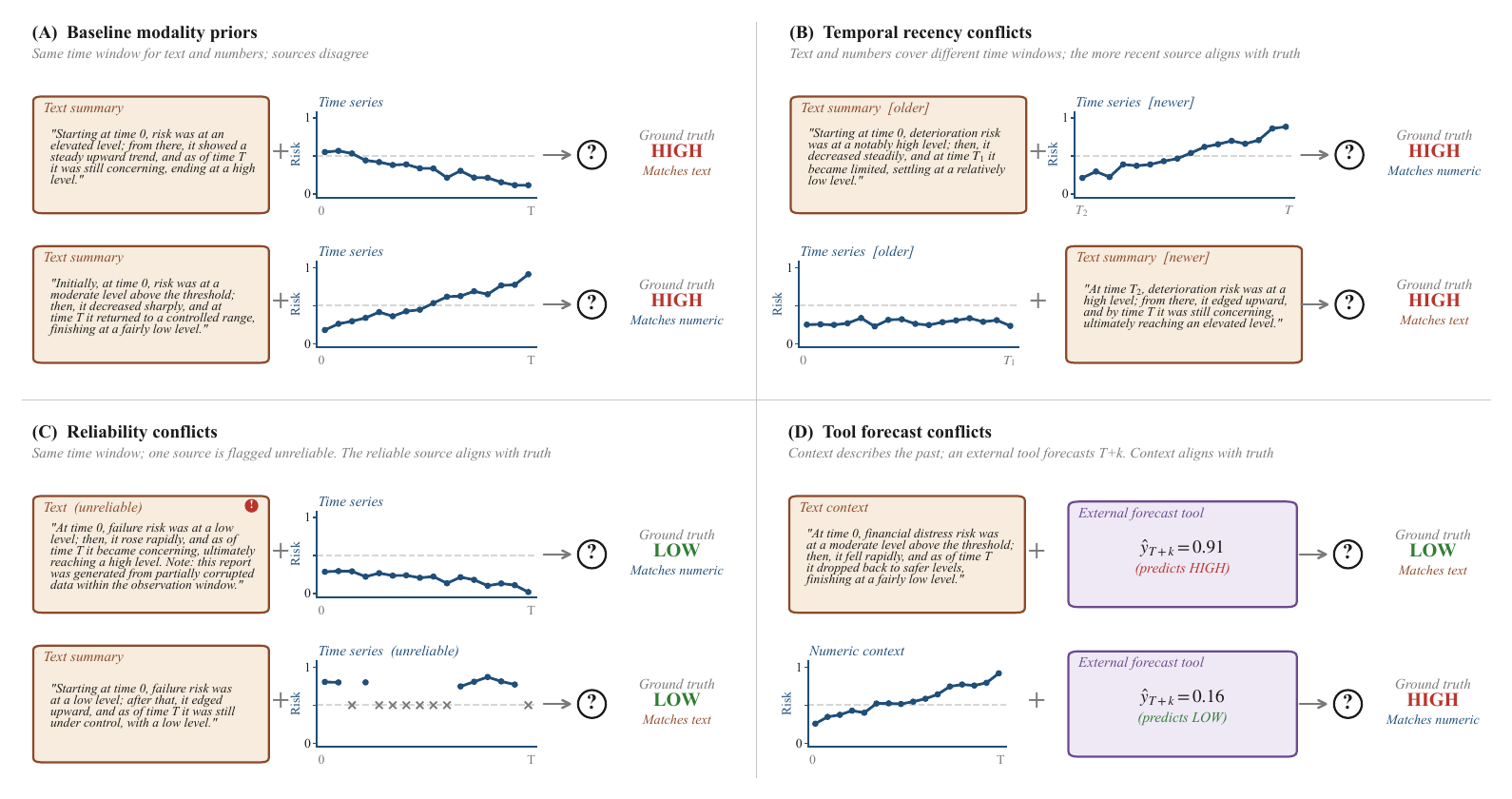}\vspace{1.0mm}
    \caption{%
        \textbf{Benchmark conflict settings for evidence arbitration.} Each panel illustrates one of the four controlled conflict settings used in our evaluation. In every setting, two evidence sources support opposing decisions and exactly one source is aligned with the ground-truth label. \textbf{(A)~Baseline modality priors:} both sources cover the same window $[0,T]$ and merely disagree. \textbf{(B)~Temporal recency:} sources cover different time windows; the more recent source is always aligned with truth. \textbf{(C)~Reliability:} one source is marked as unreliable (a corruption note in text, missing values in numbers); the reliable source is aligned with truth. \textbf{(D)~Tool forecast:} an external forecasting tool predicts a value at $T{+}k$ that contradicts the observed context; context is aligned with truth.
    }
    \label{fig:conflicts}
\end{figure*}

\subsection{Task Definition}
We study \emph{textual--numerical evidence arbitration}: the problem of deciding which source to prioritize when textual and numerical evidence supports incompatible conclusions. As illustrated in Figure~\ref{fig:pipeline}, the model receives a prompt containing two evidence sources and must answer a binary-choice question about a future target value.

Each instance is generated from a latent risk trajectory with values in $[0,1]$. The model observes evidence derived from the trajectory up to time $T$ and predicts whether the future value at $T+k$ is \textsc{high} ($>0.5$) or \textsc{low} ($<0.5$), where $k \geq 1$. Evidence is presented as a serialized numerical time series, a natural language summary, or an external forecast. By construction, only one source is aligned with the ground-truth label, while the other supports the opposite label. The model must therefore infer which source to prioritize.

We use a coarse-grained binary forecasting objective rather than precise numerical prediction, in order to isolate arbitration behaviour under conflicting evidence while reducing confounds from known limitations of LLMs in fine-grained numerical forecasting \citep{tan2024language, park2025revisiting}. 

\subsection{Conflict Dimensions}
We evaluate arbitration behaviour using four conflict settings, summarized in Figure~\ref{fig:conflicts}, which disentangle the effects of modality, temporal recency, source reliability, and evidence provenance.

\paragraph{Baseline Modality Priors.}
Textual and numerical evidence are matched in temporal scope and reliability, but support opposite labels. Across instances, the ground-truth-aligned source is alternated between text and numbers. This setting evaluates whether LLMs exhibit an inherent modality prior when resolving conflicts in the absence of additional arbitration cues.

\paragraph{Temporal Recency Conflicts.}
Textual and numerical evidence describe different temporal windows and support opposite labels. The more recent source is always aligned with the ground-truth label, and both sources are presented with explicit timestamps. This setting tests whether models use temporal recency as an arbitration cue when textual and numerical evidence disagree.

\paragraph{Reliability Conflicts.}
Textual and numerical evidence describe the same temporal window but differ in reliability. The reliable source is always aligned with the ground-truth label. For numerical evidence, unreliability is simulated by randomly masking 50\% of time-series values using \texttt{NaN} entries; for textual evidence, it is indicated through an explicit statement that the source observations are incomplete or corrupted. This setting tests whether models appropriately discount evidence marked as unreliable during arbitration.

\paragraph{Tool Forecast Conflicts.}
The model receives contextual evidence describing observations up to time $T$, in either textual or numerical form, together with a simulated external forecast for the target time $T+k$. The prompt explicitly states that the forecasting tool analyzed the same observations provided in the context before producing its prediction. By construction, however, the context is aligned with the ground-truth label, while the forecast supports the opposite label. Tool-generated forecasts are simulated as described in Appendix~\ref{sec:appendix_data_gen_tool}. This adversarial tool-conflict setting evaluates whether models over-rely on tool-generated predictions even when they conflict with contextual evidence.

\subsection{Benchmark Construction}

\begin{figure*}[t]
    \centering
    \includegraphics[width=\linewidth]{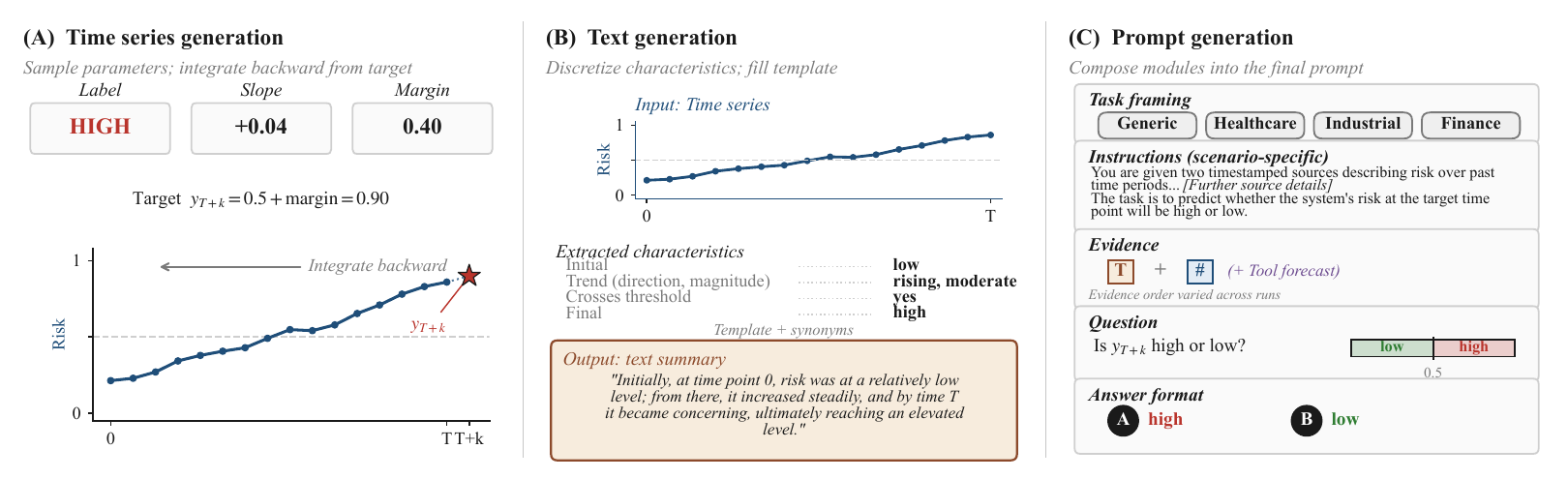}\vspace{1.0mm}
    \caption{%
        \textbf{Benchmark construction pipeline.} A single example is traced through the three stages used to generate model prompts. \textbf{(A)~Time series generation:} a sampled label, slope, and margin fix the target value $y_{T+k} = 0.5 \pm \text{margin}$, from which the observed trajectory is generated by integrating backwards with additive noise. \textbf{(B)~Text generation:} trajectory-level features are discretized and rendered as a natural language summary using templates and synonym sampling. \textbf{(C)~Prompt generation:} the textual summary and numerical series are combined with task framing, evidence-order controls, and answer choices to produce the final binary-choice prompt.
    }
    \label{fig:benchmark}
\end{figure*}

Our framework comprises three main components: a time series generator, a text generator, and a prompt generator, as illustrated in Figure~\ref{fig:benchmark}.

\paragraph{Time Series Generation.}
Latent risk trajectories follow a stochastic linear process with additive Gaussian noise and a latent slope parameter $s$ controlling the overall trend direction and magnitude:
\[
x_{t+1} = x_t + s + \epsilon_t,
\qquad
\epsilon_t \sim \mathcal{N}(0,\sigma^2).
\]
To generate each trajectory, we first sample the label, the latent slope $s$, and a margin $m$ around the decision threshold $0.5$ (Figure~\ref{fig:benchmark}A). The target value at horizon $T+k$ is then set to $0.5 + m$ for \textsc{high} and $0.5 - m$ for \textsc{low}. We generate the observed trajectory backward from step $T$ to step $1$ using the corresponding reverse-time recursion. More details about the time series generation procedure are provided in Appendix~\ref{subsec:appendix_data_gen_ts}. For the main arbitration experiments, we use observed trajectories of length $T=16$, and a forecasting horizon of $k=1$ to keep task difficulty manageable. Each value is paired with a generated timestamp, allowing the series to be serialized into a temporally grounded representation. We use a simulated sampling frequency of one minute in all experiments. For each conflict dimension, we construct balanced datasets with equal numbers of \textsc{high} and \textsc{low} labels. Examples of generated trajectories are shown in Figure~\ref{fig:ts_examples} in the appendix.

\paragraph{Text Generation.} 
The text generator converts each time series into a natural language description of its overall trajectory (Figure~\ref{fig:benchmark}B). We extract a small set of high-level characteristics from the series, including the initial level, the direction and strength of the overall trend, whether the trajectory stays on the same side of the decision threshold or transitions across it, and the final position relative to the threshold. These characteristics are discretized into semantic categories (e.g., low/moderate/high initial level, weak/moderate/strong increase or decrease, slightly or moderately above/below the threshold) and mapped to predefined natural language phrases through template-based rules. The selected phrases are then composed into complete textual summaries using randomized synonym and template choices to increase linguistic variability while preserving semantic consistency. More details about the text generation procedure are provided in Appendix~\ref{subsec:appendix_data_gen_text}. To maintain a clear separation between textual and numerical evidence, the generated descriptions never include explicit numerical values or exact measurements from the underlying time series. 

To create conflicts between textual and numerical evidence, we keep the original numerical evidence unchanged and sample a second trajectory conditioned on the opposite label. This newly generated series is then converted into text using the same generation pipeline, yielding textual evidence that is semantically coherent but inconsistent with the accompanying numerical evidence.

\paragraph{Prompt Generation.}
Prompts are constructed in a modular fashion (Figure~\ref{fig:benchmark}C). The first component introduces the task and optionally provides domain-specific framing. We consider a generic framing, along with three domain-specific instantiations: healthcare, industrial, and finance. We then present the available evidence, which may include textual summaries, numerical time series observations, and/or simulated external forecast predictions. After presenting the evidence, the prompt explicitly asks the model to predict whether the target risk value will be \textsc{high} or \textsc{low}. The prediction is formulated as a binary-choice decision between answer options ``A'' and ``B''. We provide examples of full prompts in Appendix~\ref{sec:appendix_prompt_examples}. We systematically control the order in which evidence sources are presented (i.e., whether the source aligned with the ground truth appears first or last in the prompt). This manipulation allows us to study how evidence presentation order influences arbitration behaviour. Additional details on prompt design can be found in Appendix~\ref{sec:appendix_prompts}.

\subsection{Evaluation Protocol}
We evaluate arbitration behaviour using prompt templates tailored to each experiment and applied consistently across all models. For each sample, the model selects one of the two answer options (``A'' or ``B'') based on the provided textual and/or numerical evidence. Predictions are then obtained directly from the logits associated with the binary answer tokens, which are mapped to the corresponding labels.

For each experiment, we report classification accuracy under conflicting evidence conditions, alongside unimodal reference conditions in which only the evidence source aligned with the ground truth is provided. Unless otherwise specified, all experiments use the default configuration shown in Table~\ref{tab:default_config}. Results are averaged across three random seeds.

\subsection{Models}
We evaluate a diverse set of open-weight instruction-tuned LLMs spanning multiple architectures and parameter scales, including the \qwenfamily family (1.7B, 4B, 8B, and 14B) \citep{yang2025qwen3}, \gemmafull (\gemma) \citep{gemmateam2024gemma2improvingopen}, \llamafull (\llama) \citep{grattafiori2024llama}, and \mistralfull (\mistral) \citep{Jiang2023Mistral7}. This selection enables comparisons both across model families and within a controlled scaling series.

\section{Results}
\label{sec:results}
We evaluate arbitration behaviour across a range of controlled conflict settings designed to isolate the effects of modality, temporal recency, source reliability, and evidence provenance, as well as through sensitivity analyses examining the impact of domain instantiations and answer choice configurations. For the main arbitration experiments, we generate balanced datasets containing 2000 instances per setting (1000 per label), while sensitivity analyses use 1000 instances per setting for computational efficiency.

\begin{figure*}[t]
    \centering

    \begin{subfigure}{0.48\linewidth}
        \centering
        \includegraphics[width=\linewidth]{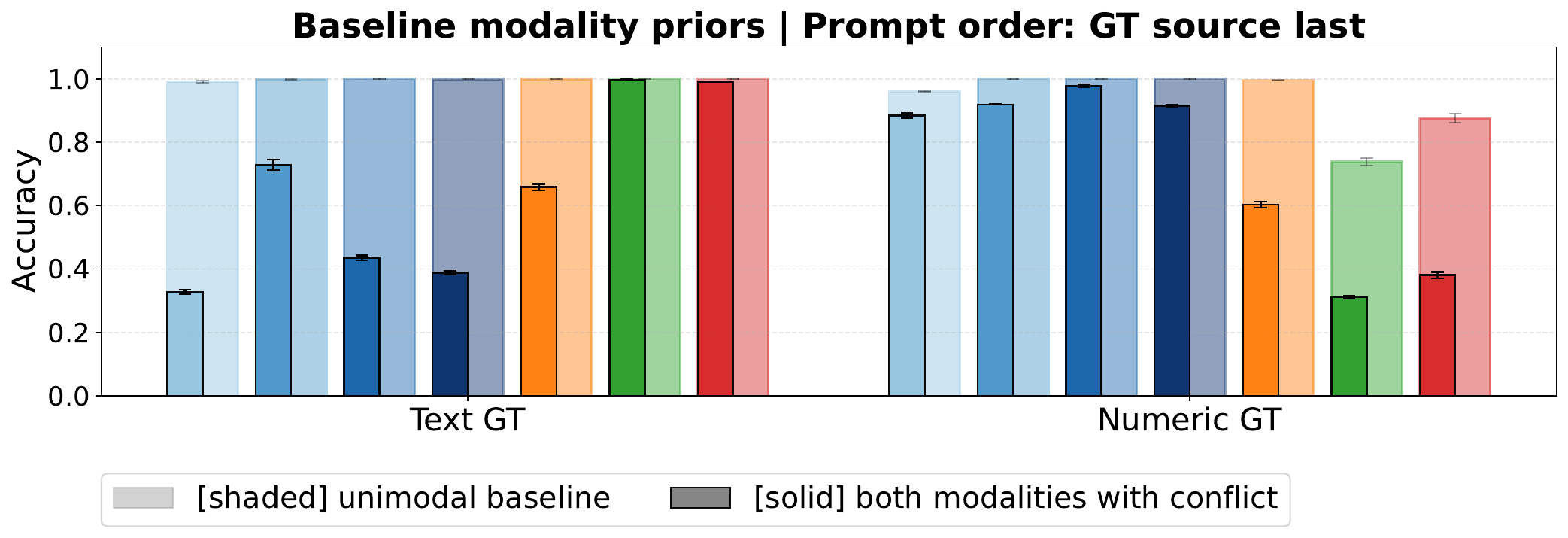}\vspace{-2.0mm}
        \caption{}
        \label{fig:main_results_a}
    \end{subfigure}
    \hfill
    \begin{subfigure}{0.48\linewidth}
        \centering
        \includegraphics[width=\linewidth]{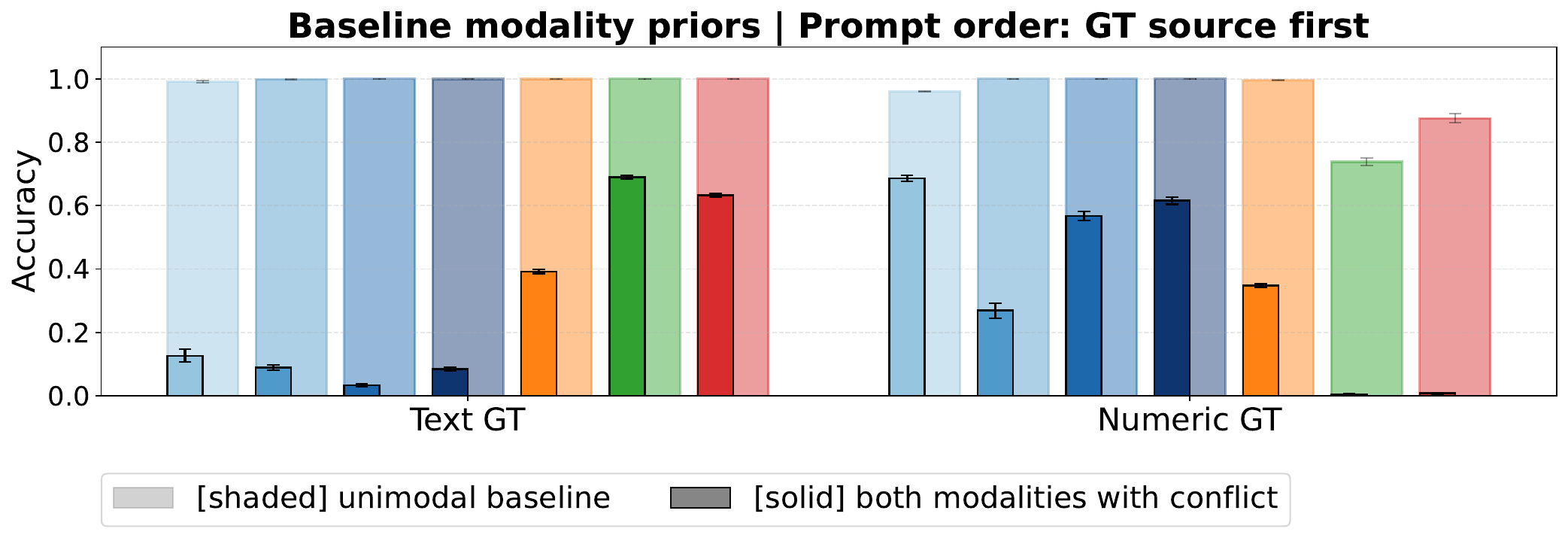}\vspace{-2.0mm}
        \caption{}
        \label{fig:main_results_b}
    \end{subfigure}

    \vspace{0.30em}

    \begin{subfigure}{0.48\linewidth}
        \centering
        \includegraphics[width=\linewidth]{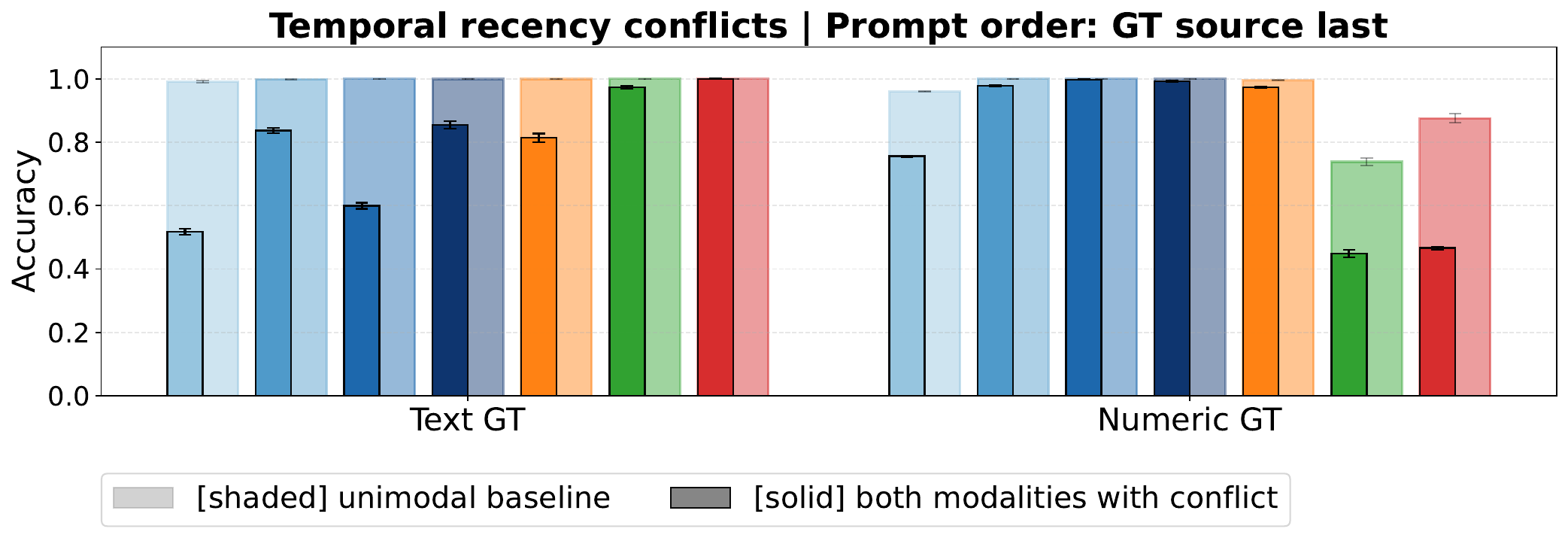}\vspace{-2.0mm}
        \caption{}
        \label{fig:main_results_c}
    \end{subfigure}
    \hfill
    \begin{subfigure}{0.48\linewidth}
        \centering
        \includegraphics[width=\linewidth]{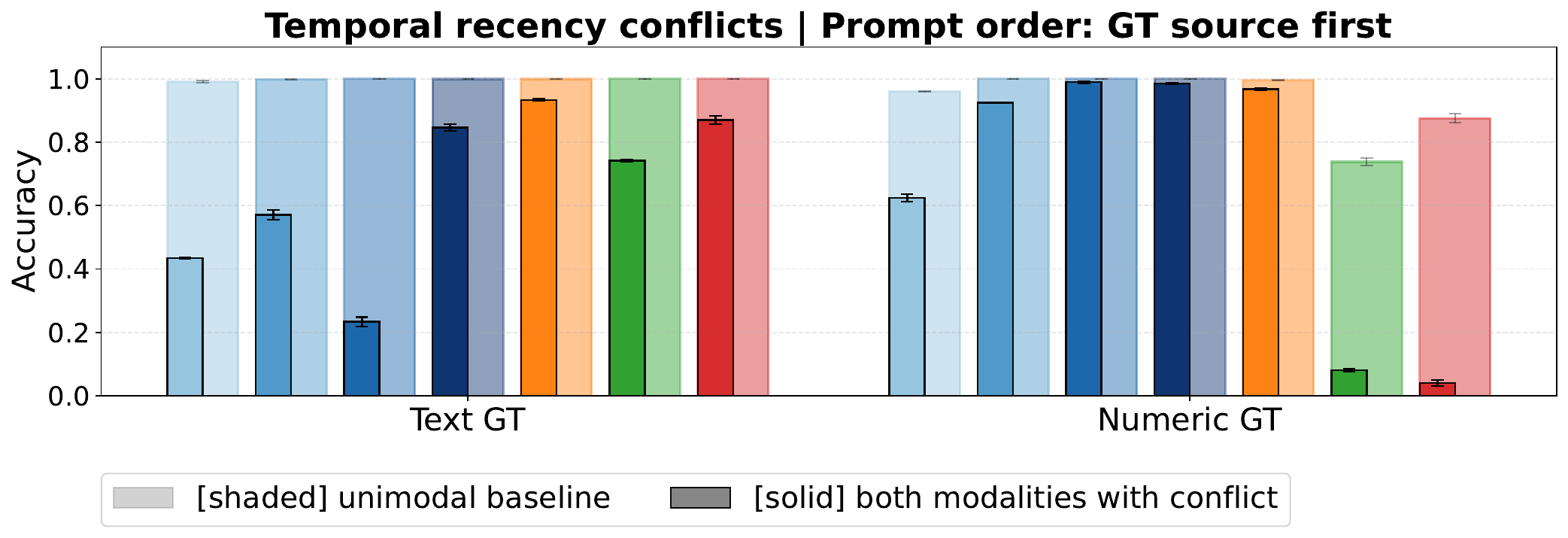}\vspace{-2.0mm}
        \caption{}
        \label{fig:main_results_d}
    \end{subfigure}

    \vspace{0.30em}
    
    \begin{subfigure}{0.48\linewidth}
        \centering
        \includegraphics[width=\linewidth]{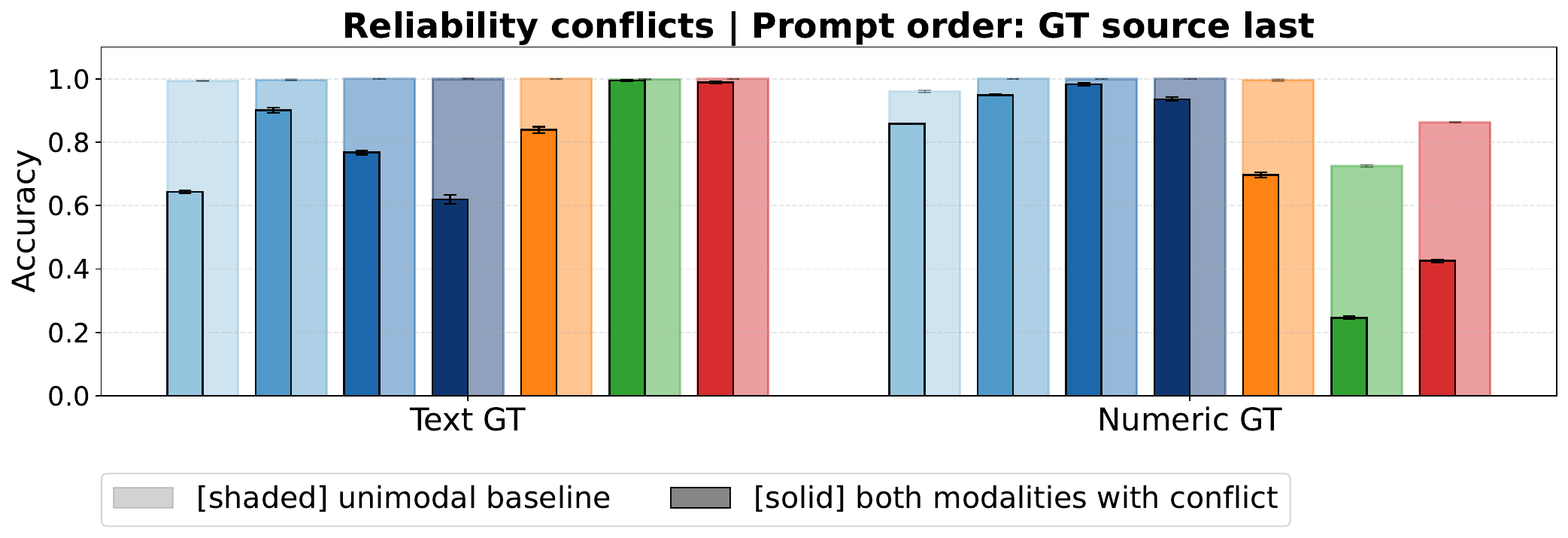}\vspace{-2.0mm}
        \caption{}
        \label{fig:main_results_e}
    \end{subfigure}
    \hfill
    \begin{subfigure}{0.48\linewidth}
        \centering
        \includegraphics[width=\linewidth]{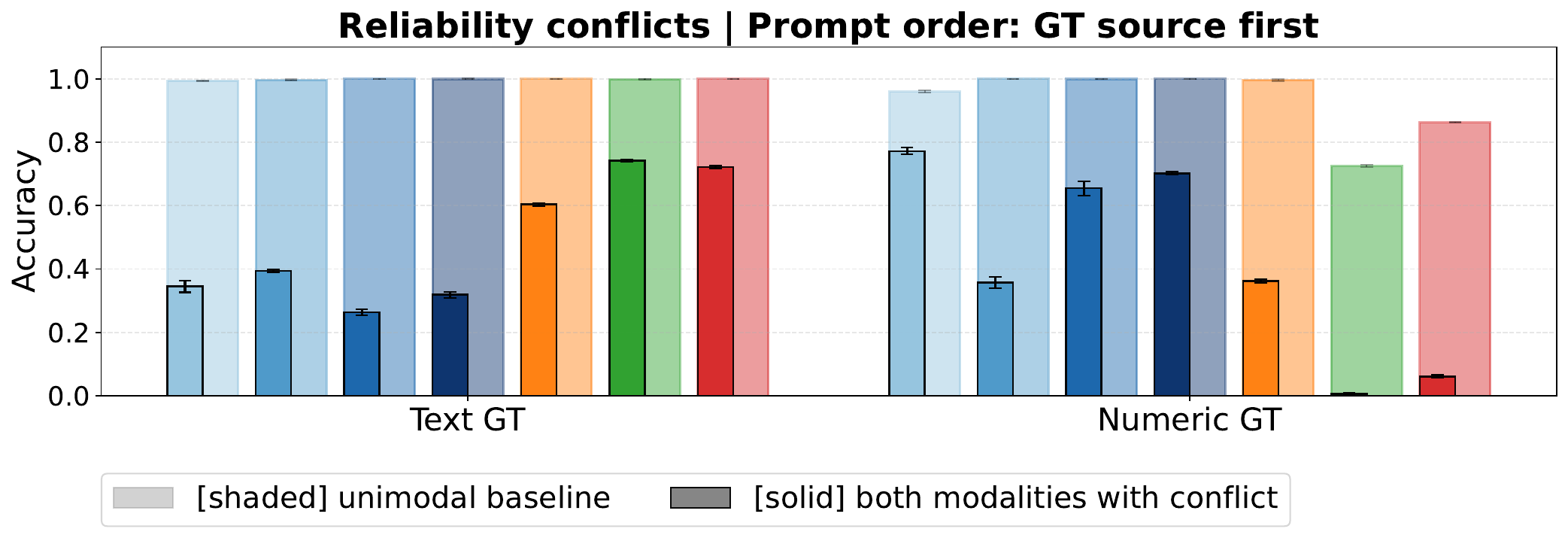}\vspace{-2.0mm}
        \caption{}
        \label{fig:main_results_f}
    \end{subfigure}

    \vspace{0.30em}

    \begin{subfigure}{0.48\linewidth}
        \centering
        \includegraphics[width=\linewidth]{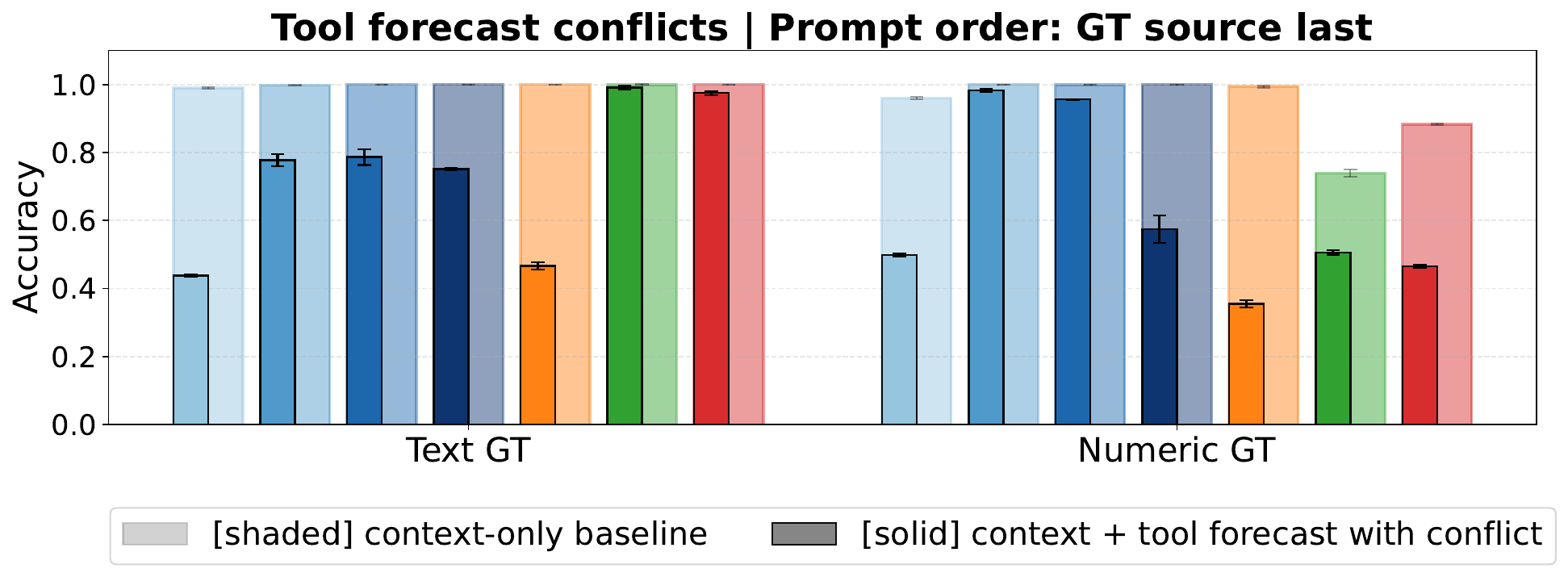}\vspace{-2.0mm}
        \caption{}
        \label{fig:main_results_g}
    \end{subfigure}
    \hfill
    \begin{subfigure}{0.48\linewidth}
        \centering
        \includegraphics[width=\linewidth]{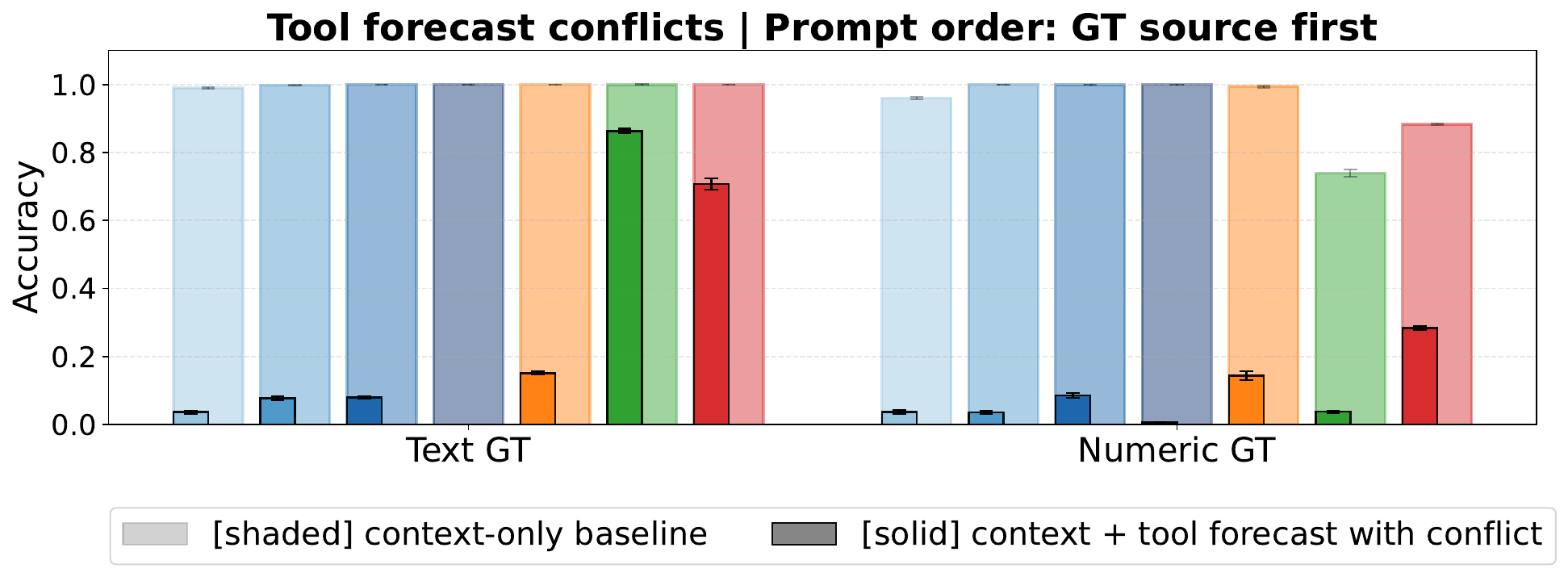}\vspace{-2.0mm}
        \caption{}
        \label{fig:main_results_h}
    \end{subfigure}

    \vspace{0.05em}

    \begin{subfigure}{0.96\linewidth}
        \centering
        \includegraphics[width=\linewidth]{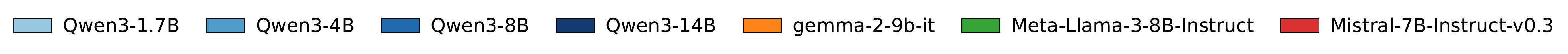}
    \end{subfigure}

    \vspace{1.0mm}
    \caption{\textbf{Arbitration accuracy under conflicting evidence.} Panels show classification accuracy across the four conflict settings and evidence-order conditions: \textbf{(A--B)} baseline modality priors, \textbf{(C--D)} temporal recency, \textbf{(E--F)} reliability, and \textbf{(G--H)} tool forecast conflicts. In each setting, exactly one evidence source is aligned with the ground-truth label (GT). Shaded bars show unimodal reference accuracy using only the ground-truth-aligned source, while solid bars show accuracy when both conflicting sources are provided. Results are averaged over three random seeds.}
    \label{fig:main_results}
\end{figure*}

\subsection{Baseline Modality Priors}
\label{subsec:results_modality_priors}
Results in Figures~\ref{fig:main_results_a}--\ref{fig:main_results_b} show that all models achieve very high unimodal accuracy, indicating that performance differences in the conflicting setting primarily reflect arbitration behaviour rather than intrinsic task difficulty. Distinct modality priors emerge across model families. \qwenfamily models consistently favour numerical evidence, whereas \llama and \mistral models show comparatively stronger reliance on textual evidence. \gemma exhibits the most balanced behaviour between the two modalities.

The numerical preference of \qwenfamily is particularly pronounced: all \qwenfamily variants systematically favour numerical evidence even when it conflicts with perfectly predictive textual context. No clear monotonic relationship with model size is observed. In contrast, the complementary behaviour of \llama and \mistral may partially reflect weaker overall capability to interpret the numerical evidence, as these models also obtain slightly lower unimodal numerical-only accuracies compared to \qwenfamily and \gemma.

Evidence order substantially affects arbitration behaviour across nearly all models. Accuracy is generally higher when the source aligned with the ground truth appears later in the prompt, revealing a strong prompt recency effect. This effect interacts with modality priors: numerical evidence remains influential regardless of position, whereas textual evidence benefits much more strongly from appearing last. In several conditions, larger \qwenfamily variants even fall below chance (accuracy below $0.5$) when textual evidence is correct and numerical evidence conflicts, suggesting a systematic bias toward numerical evidence rather than simple uncertainty.

\subsection{Temporal Recency Conflicts}
\label{subsec:results_recency}
As shown in Figures~\ref{fig:main_results_c}--\ref{fig:main_results_d}, compared to the baseline modality prior setting, the temporal recency setting produces much stronger and more consistent arbitration behaviour across model families, suggesting that temporal recency is a particularly influential cue for resolving conflicting evidence. We observe that the order of evidence presentation again has a substantial effect: performance is generally higher when the most recent source is presented later in the prompt, reinforcing the prompt recency effects already observed in the baseline experiments. Nevertheless, models differ in their sensitivity to evidence order.

\gemma exhibits the most consistent behaviour, maintaining high accuracy regardless of evidence order. \qwenfamily models also follow temporal recency cues very reliably, although some variants are more sensitive to prompt ordering than \gemma. As in the baseline experiments, no clear monotonic relationship with model size is observed.

\subsection{Reliability Conflicts}
\label{subsec:results_reliability}
Across both evidence order settings (Figures~\ref{fig:main_results_e}--\ref{fig:main_results_f}), reliability conflicts produce larger performance drops than temporal recency conflicts across most models, suggesting that explicit source reliability is a weaker arbitration cue than temporal recency.
\gemma again exhibits the most stable behaviour across ordering conditions, whereas \qwenfamily variants appear more sensitive to evidence order despite achieving some of the highest peak accuracies.

\subsection{Tool Forecast Conflicts}
\label{subsec:results_tool}
Figures~\ref{fig:main_results_g}--\ref{fig:main_results_h} show that tool forecast conflicts produce the strongest degradation observed across all experiments, indicating that many models heavily over-rely on external forecasts even when these systematically conflict with the provided contextual measurements. 

Evidence order has a particularly strong effect in this setting. Accuracy improves substantially when the contextual evidence is presented after the tool forecast, indicating that later evidence can partially mitigate over-reliance on the external prediction. Relative to the baseline modality-prior experiments, the introduction of an explicit forecast greatly amplifies arbitration failures, especially for \qwenfamily and \gemma. These models are particularly susceptible when the ground-truth-aligned contextual evidence appears first, often achieving near-zero accuracy despite perfectly predictive contextual evidence.

\llama and \mistral are substantially less influenced by incorrect tool forecasts, frequently retaining relatively high accuracy even in the conflicting setting. 

\subsection{Sensitivity Analysis}
\label{subsec:results_sensitivity}
To assess robustness to domain instantiation and answer choice configuration, we perform sensitivity analyses in the same setting used for the baseline modality prior experiments. We vary either the domain or the answer choice configuration, while keeping all remaining parameters fixed to the default hyperparameter values in Table~\ref{tab:default_config} and using the default domain (i.e., generic), label semantics (``A''=\textsc{high}), and answer ordering (i.e., ``A'' first) as the baseline configuration. For each sweep condition, we compute accuracy differences relative to this baseline and report the absolute deltas aggregated across sweep values, seeds and evidence-order settings as mean $\pm$ standard deviation, separately for each model. 

As shown in Table~\ref{tab:sensitivity}, sensitivity to both domain specialization and answer choice configuration is generally low in text-only settings, but more noticeable effects emerge in numeric-only and conflicting settings for several models. In particular, answer choice perturbations can produce significant shifts in conflict accuracy despite relatively stable unimodal performance, indicating that arbitration behaviour can depend on superficial prompt structure. Robustness also tends to improve with scale within the \qwenfamily family, with the 14B model remaining comparatively stable across all settings. Additional sensitivity analyses on data-generation parameters are described in Appendix~\ref{sec:appendix_sensitivity}.

\section{Discussion}
Across all experiments, arbitration behaviour is highly systematic rather than random. Models consistently rely on salient evidence characteristics, including modality, temporal recency, source reliability, and external forecasts, even when these cues conflict with the ground truth. Temporal recency emerges as the most consistently followed arbitration signal across model families, whereas reliability cues are weaker and lead to substantially larger performance degradation. External forecasts are particularly influential: tool forecast conflicts produce the strongest failures overall, indicating that many models heavily privilege explicit predictions over directly observed contextual measurements.

Distinct arbitration patterns also emerge across model families. \qwenfamily models consistently favour numerical evidence and are especially susceptible to misleading external forecasts, while \llama and \mistral rely comparatively more on textual evidence, albeit with overall lower accuracy. \gemma exhibits the most balanced and stable behaviour across settings. Importantly, these behaviours do not scale monotonically with model size within the \qwenfamily family, suggesting that arbitration biases are not simply a function of parameter count.

Evidence presentation order plays a major role in arbitration. Across all experimental settings, evidence presented later in the prompt tends to exert greater influence on the final prediction, partially overriding earlier conflicting information. This positional effect often amplifies the underlying arbitration cue itself, for example strengthening the influence of temporally recent evidence or external forecasts when they appear last. Sensitivity analyses further show that arbitration behaviour also can vary under changes in answer choice configuration and domain framing, although these effects are generally smaller.

Several settings produce below-chance or near-zero accuracy. This indicates that models are not merely uncertain under conflict, but can systematically favour incorrect evidence sources. 
Overall, the results suggest that current LLMs rely heavily on heuristic arbitration strategies rather than robust evidence integration, making them vulnerable to predictable and systematic failures in multi-source decision-making settings.

\section{Conclusion}
As LLMs are increasingly embedded in decision-making pipelines, their ability to handle conflicting evidence becomes central to their reliability. This work studied evidence arbitration between textual summaries, numerical observations, and external tool outputs that support incompatible conclusions, by introducing a controlled synthetic benchmark that isolates key arbitration cues (modality, temporal recency, source reliability, and evidence provenance).

Our results show that LLMs do not resolve such conflicts randomly. Instead, they exhibit systematic, model-specific arbitration patterns, often relying on heuristic cues when deciding which evidence to trust. Temporal recency is followed more consistently than explicit reliability information, while external forecasts can exert disproportionate influence even when they conflict with direct contextual evidence. 

These findings suggest that evaluating LLMs on isolated textual, numerical, or tool-use tasks is insufficient for understanding their behaviour in multi-source decision settings. Conflict-based evaluations provide a useful stress test for evidence integration, and we hope this benchmark motivates further work on arbitration under real-world source conflicts in tool-augmented systems.

\section*{Limitations}
This work does not include experiments on real-world data. Instead, the synthetic framework intentionally simplifies real-world decision-making settings in order to provide full control over arbitration cues, including temporal recency, source reliability, and evidence provenance, enabling systematic analysis of arbitration behaviour under conflict.

In addition, our task formulation casts forecasting as a binary decision problem, which does not capture the full complexity of numerical forecasting tasks. However, this design reduces confounds arising from known limitations of current LLMs in accurate numerical prediction, allowing cleaner evaluation of how models prioritize conflicting evidence sources.

\paragraph{Potential Risks.} Our benchmark is intended solely for evaluating evidence arbitration in LLMs and should not be interpreted as guidance for deploying such models in high-stakes decision-making settings.

\section*{Acknowledgments}
DAC was funded by an NIHR Research Professorship (NIHR302440); a Royal Academy of Engineering Research Chair; and the InnoHK Hong Kong Centre for Cerebro-cardiovascular Engineering (COCHE); and was supported by the National Institute for Health Research (NIHR) Oxford Biomedical Research Centre (BRC) and the Pandemic Sciences Institute at the University of Oxford.

\bibliographystyle{unsrtnat}
\bibliography{references} 

\clearpage
\appendix

\renewcommand{\thefigure}{A\arabic{figure}}
\setcounter{figure}{0}

\renewcommand{\thetable}{A\arabic{table}}
\setcounter{table}{0}

\renewcommand{\theequation}{A\arabic{equation}}
\setcounter{equation}{0}

\section{Use of AI Assistants}
AI assistants were used only to improve the phrasing, clarity, and grammar of this manuscript. All AI-generated text was reviewed and revised by the authors, who take full responsibility for the final content.

\section{Data Generation Details}
\label{sec:appendix_data_gen}
In this section, we provide implementation details for the synthetic data generation pipeline used throughout the experiments.

\subsection{Time Series Generation}
\label{subsec:appendix_data_gen_ts}
Trajectories are generated using a stochastic linear process with additive Gaussian noise and a latent slope parameter controlling the overall trend direction and magnitude. For all main experiments, we use trajectories of length $T=16$ and a forecasting horizon of $k=1$. In additional sensitivity analyses, we vary the trajectory length using $T \in \{8,16,32\}$. The target value at time point $T+k$ is first sampled according to the desired class label. Let $m$ denote the sampled margin from the decision threshold $0.5$:
\[
m \sim \mathcal{U}(m_{\text{low}}, m_{\text{high}}),
\]
where $m_{\text{low}}=0.35$ and $m_{\text{high}}=0.45$ in the main experiments. In additional sensitivity analyses, we vary the margin range using $(m_{\text{low}}, m_{\text{high}}) \in \{(0.15, 0.25), (0.25, 0.35), (0.35, 0.45)\}$. The future target value is then defined as
\[
y_{T+k} =
\begin{cases}
0.5 + m & \text{if label = \textsc{high}}, \\
0.5 - m & \text{if label = \textsc{low}}.
\end{cases}
\]

To avoid degenerate near-random forecasting instances, we require the minimum margin to dominate the cumulative stochastic noise over the forecasting horizon. Specifically, generation is rejected whenever
\[
m_{\text{low}} < 1.5 \cdot \sigma \sqrt{k},
\]
where $\sigma$ denotes the standard deviation of the Gaussian noise process.

For each trajectory, a latent slope parameter is independently sampled from a symmetric uniform distribution
\[
s \sim \mathcal{U}(-0.05, 0.05),
\]
where the range of the uniform distribution is chosen to minimize rejection of the generated trajectories due to violations of the valid risk range $[0,1]$.
Importantly, the slope distribution is shared across labels to prevent shortcut correlations between slope and class membership.

Given the sampled future target value $y_{T+k}$, we first estimate the final observed value $x_T$ by approximately inverting the forward stochastic process
\[
x_T = y_{T+k} - s \cdot k - \epsilon,
\]
where
\[
\epsilon \sim \mathcal{N}(0, \sigma^2)
\]
and $\sigma = 0.05$ in the main experiments. In additional sensitivity analyses, we vary the noise level using $\sigma \in \{0.01, 0.05, 0.1\}$. Starting from this estimated value $x_T$, the observed trajectory is then generated backward in time from step $T$ to step $1$. Let $x_t$ denote the trajectory value at time step $t$. The backward dynamics follows
\[
x_t = x_{t+1} - s + \epsilon_t, \quad \epsilon_t \sim \mathcal{N}(0, \sigma^2).
\]
A forward simulation step is then performed to verify that the resulting future target remains consistent with the desired label and sampled margin constraint. Trajectories violating the target label, margin condition, or valid risk range $[0,1]$ are rejected and regenerated using rejection sampling.
Across all experiments, the rejection sampling procedure converged reliably within a small number of attempts.

Each trajectory value is associated with a synthetic timestamp sampled at one-minute intervals. For each instance, a random starting hour is uniformly sampled between 08:00 and 16:00, together with a random starting minute. Timestamps are then generated sequentially using a fixed simulated sampling frequency of one minute.

For every conflict dimension, datasets are constructed to maintain balanced class distributions, with equal numbers of \textsc{high} and \textsc{low} labels. 

\begin{figure}[t]
  \centering
  \includegraphics[width=0.75\columnwidth]{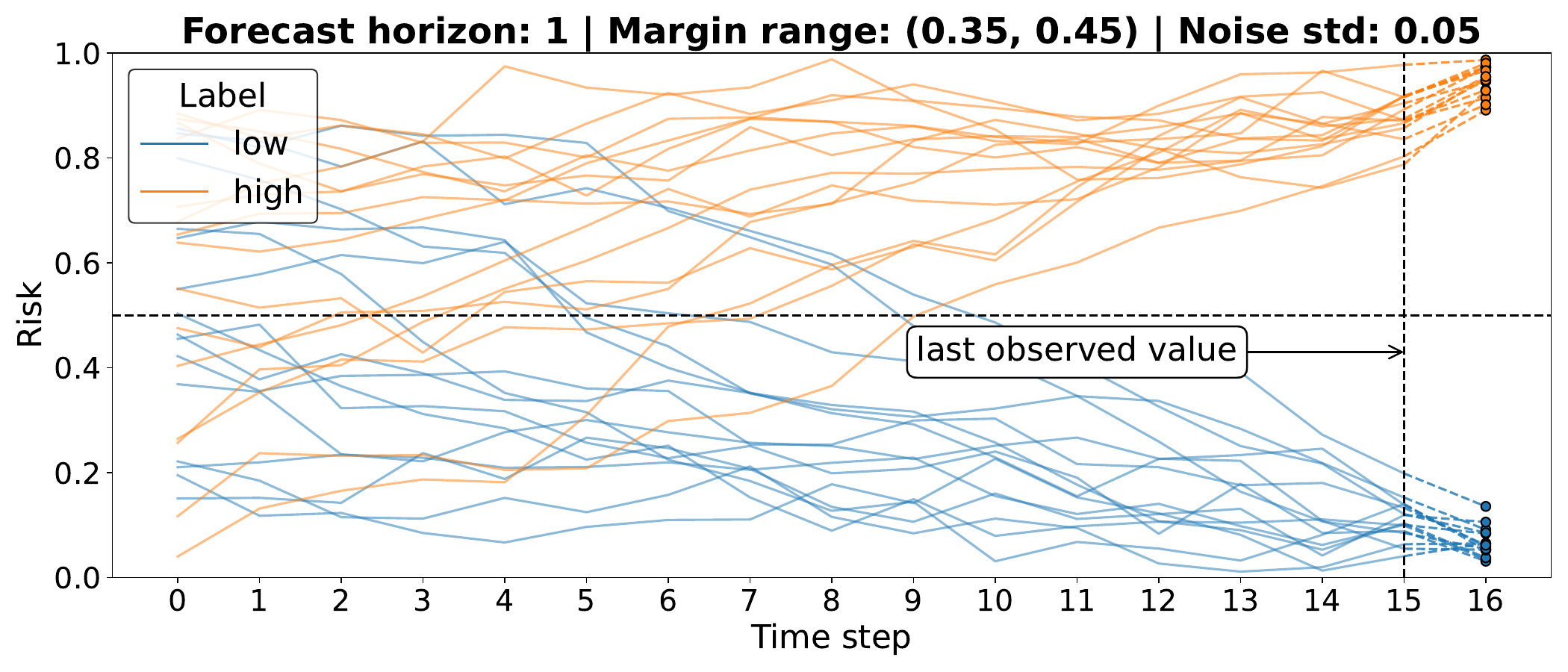}\vspace{1.0mm}
  \caption{%
    \textbf{Representative examples of generated time series.} Generated trajectories with observable length $T=16$ and forecast horizon $k=1$.
  }
  \label{fig:ts_examples}
\end{figure}

Examples of generated trajectories are shown in Figure~\ref{fig:ts_examples}.

\subsection{Text Generation}
\label{subsec:appendix_data_gen_text}

Each time series is summarized through a small set of global trajectory characteristics extracted from the full sequence. The generated text is intentionally high level: it describes coarse temporal behaviour while avoiding explicit numerical values or exact measurements. This preserves a clear separation between textual and numerical evidence modalities. All generated textual summaries are produced in English using a template-based generation procedure with randomized lexical variation.

\paragraph{Extracted trajectory features.}
For each time series, we extract the initial observed value, the final observed value, the overall linear trend direction and magnitude, and whether the trajectory remains on the same side of the decision threshold or crosses it. The decision threshold is fixed at $0.5$. Descriptions are therefore expressed relative to this threshold (e.g., ``below the threshold'' or ``above the critical level'').

\paragraph{Level discretization.}
Continuous values are discretized into semantic categories before text generation. The following bins are used:

\begin{center}
\begin{tabular}{ll}
\toprule
Level category & Value range \\
\midrule
\texttt{low} & $x < 0.30$ \\
\texttt{mid\_below} & $0.30 \leq x < 0.45$ \\
\texttt{slightly\_below} & $0.45 \leq x < 0.50$ \\
\texttt{slightly\_above} & $0.50 \leq x \leq 0.55$ \\
\texttt{mid\_above} & $0.55 < x \leq 0.70$ \\
\texttt{high} & $x > 0.70$ \\
\bottomrule
\end{tabular}
\end{center}

Each category is mapped to multiple synonymous natural language realizations. For example, \texttt{slightly\_below} may be rendered as ``a level slightly below the threshold'' or ``a value marginally below the critical level''.

\paragraph{Trend discretization.}
The global trend is computed from the scalar slope associated with the generated time series. Trend magnitude is discretized into three categories according to the absolute slope value:

\begin{center}
\begin{tabular}{ll}
\toprule
Magnitude & Condition \\
\midrule
\texttt{weak} & $|s| < 0.015$ \\
\texttt{moderate} & $0.015 \leq |s| < 0.03$ \\
\texttt{strong} & $|s| \geq 0.03$ \\
\bottomrule
\end{tabular}
\end{center}

Trend direction is determined by the sign of the slope. Each direction-strength pair is mapped to multiple paraphrased textual descriptions (e.g., ``rose gradually'', ``increased steadily'', ``climbed significantly'').

\paragraph{Cross-threshold trajectory semantics.}
The final sentence of each generated description depends jointly on the initial and final regions relative to the threshold. Four cases are distinguished: remaining below the threshold, remaining above the threshold, crossing from below to above, crossing from above to below. For example, trajectories crossing from below to above may yield phrases such as ``became concerning'' or ``rose into an elevated state'', whereas trajectories remaining below threshold may produce ``continued to stay contained'' or ``still remained limited''.

\paragraph{Template composition and lexical variation.}
Text generation follows a template-based pipeline with randomized lexical choices. Independent synonym banks are used for:
\begin{itemize}
    \item sentence opening expressions introducing the initial timestamp (e.g., ``Initially, at 12:48, \dots'', ``Starting at 14:17, \dots''),
    \item temporal closing expressions referring to the final timestamp (e.g., ``\dots by 13:52'', ``\dots as of 09:15''),
    \item synonymous references to the decision threshold (e.g., ``the threshold'', ``the critical level'', ``the cutoff level''),
    \item discourse connectors linking trajectory stages (e.g., ``then'', ``after that'', ``from there''),
    \item descriptions of initial and final value ranges relative to the threshold (e.g., ``a moderate value below the threshold'', ``an elevated level''),
    \item descriptions of trajectory evolution and final status (e.g., ``continued to stay contained'', ``moved into a concerning range'', ``returned to a controlled range'').
\end{itemize}
Random sampling from these synonym sets increases linguistic diversity while preserving the underlying semantic content. Importantly, randomness only affects lexical realization and not the semantics associated with a given trajectory.

\paragraph{Domain conditioning.}
The generation procedure is identical across domains. Only the domain-specific risk term changes. Specifically, we use:
\begin{itemize}
    \item ``risk'' for the generic domain;
    \item ``deterioration risk'' for the healthcare domain;
    \item ``failure risk'' for the industrial domain;
    \item ``financial distress risk'' for the finance domain.
\end{itemize}
The entity reference (``system'', ``patient'', ``machine'', and ``asset'', respectively) is omitted from the generated summaries, as it is already specified in the instruction component of the prompt and repeating it would introduce unnecessary redundancy.

\paragraph{Unreliability cues.} We inject unreliability cues into the generated text by appending an explicit statement indicating that the underlying observations on which the summary is based are incomplete or corrupted. Sample unreliability statements include ``Note: this report was generated from partially corrupted data within the observation window'' and ``Note: this summary was produced using partially corrupted data from the observation window''.

\paragraph{Abstraction gap between modalities.}
The textual modality intentionally summarizes only global trajectory properties and omits local fluctuations, short-term oscillations, and exact magnitudes present in the numerical series. As a result, the textual evidence represents an abstract semantic interpretation of the underlying time series rather than a verbalization of every datapoint.

\paragraph{Examples.}
Figure~\ref{fig:conflicts} presents representative examples of generated textual evidence across different domains.

\subsection{Tool Forecast Generation}
\label{sec:appendix_data_gen_tool}
Simulated tool forecasts are generated as intentionally incorrect risk predictions. For each sample, a scalar risk value is sampled from the opposite side of the decision threshold $0.5$ relative to the true label. We additionally control how confidently incorrect the tool prediction is by enforcing a minimum distance from the decision threshold (set to $0.15$ in all our experiments), preventing ambiguous forecasts close to $0.5$.

\section{Prompts}
\label{sec:appendix_prompts}
All prompts are generated programmatically using a modular template-based framework. Each prompt is composed of the following components:
\begin{enumerate}
    \item Domain framing
    \item Task instructions
    \item Evidence blocks
    \item Prediction question
    \item Answer choices
    \item Closing instruction
\end{enumerate}
This modular design enables controlled manipulation of domain instantiation, evidence modality, evidence ordering, and answer ordering while keeping the overall prompt structure fixed across experiments.

\subsection{Domain Framing}

Each prompt begins with a short domain-specific framing describing the meaning of the risk variable. Following the text generation setup, we consider one generic framing and three domain-specific instantiations (healthcare, industrial, and finance). For example, the healthcare framing is:
\begin{lstlisting}
The patient has a deterioration risk between 0 and 1, where higher values indicate greater risk of clinical deterioration and lower values indicate lower risk.
\end{lstlisting}
The underlying prediction task remains identical across domains, only the semantic framing changes.

\subsection{Task Instructions}

The instruction block depends on the available evidence modalities. We consider five task configurations: numerical evidence only, textual evidence only, textual and numerical evidence, numerical evidence with an external tool forecast, and textual evidence with an external tool forecast.

For unimodal settings, prompts state
\begin{lstlisting}
You are given a time series of numerical risk observations over a past time period.
\end{lstlisting}
for numerical evidence only, and
\begin{lstlisting}
You are given a text summary of risk observations over a past time period.
\end{lstlisting}
for textual evidence only.

For multimodal settings, prompts explicitly specify whether the two evidence sources refer to the same or different temporal windows. For same-window settings, prompts state
\begin{lstlisting}
You are given two sources describing risk over the same time window: a text summary of risk observations and a time series of numerical risk observations.
\end{lstlisting}
For temporal recency settings, prompts instead state
\begin{lstlisting}
You are given two timestamped sources describing risk over past time periods: a text summary of risk observations and a time series of numerical risk observations. The two sources may refer to different time windows, and one source may be earlier or more recent than the other. 
\end{lstlisting}

Finally, in settings involving external tool forecasts, prompts extend the unimodal instructions with an additional clause describing the tool prediction:
\begin{lstlisting}
You are given <unimodal source>. An external forecasting tool has analyzed these same observations and produced a predicted risk value at a future time point.
\end{lstlisting}

\subsection{Evidence Formatting}

\paragraph{Numerical evidence.}
Numerical evidence is presented as timestamp--value pairs:
\begin{lstlisting}
Time series:
12:38: 0.38
12:39: 0.41
12:40: 0.48
12:41: 0.47
...
\end{lstlisting}
All numerical values are displayed with two decimal places.

\paragraph{Textual evidence.}
Textual evidence is presented as a quoted natural language summary:

\begin{lstlisting}
Text summary:
"Starting at 13:10, risk was at a value marginally below the cutoff level; after that, it fell at a moderate pace, and as of 13:25 it continued to stay contained, settling at a low level."
\end{lstlisting}

\paragraph{External tool forecasts.}
External forecasts are formatted as scalar predictions associated with the target timestamp:
\begin{lstlisting}
External forecasting tool risk prediction at time 13:01: 0.76
\end{lstlisting}

\subsection{Evidence Order Manipulation}

To study ordering effects, the relative ordering of evidence blocks is systematically varied across experiments.

\paragraph{Baseline modality-prior experiments.}
We vary whether the ground-truth-aligned or ground-truth-misaligned modality appears first.

\paragraph{Temporal recency experiments.}
We vary whether the more recent or less recent source appears first.

\paragraph{Reliability experiments.}
We vary whether the more reliable or less reliable source appears first.

\paragraph{Tool forecast experiments.}
We vary whether the primary context source or the external tool forecast appears first.

\subsection{Prediction Question and Answer Choices}

After presenting the evidence, prompts ask the model to predict whether the target risk value will exceed a threshold of $0.5$:
\begin{lstlisting}
Question: Based on the information above, will the system's risk at time 15:33 be high (> 0.5) or low (< 0.5)?
\end{lstlisting}
Predictions are formulated as binary-choice decisions using answer options ``A'' and ``B''. We systematically vary the order in which answer choices appear and the mapping between answer tokens and labels. For example:
\begin{lstlisting}
A) high
B) low
\end{lstlisting}
or
\begin{lstlisting}
B) high
A) low
\end{lstlisting}
This controls for potential positional or token-level biases.

\subsection{Closing Instruction}
Each prompt concludes with a strict response constraint:
\begin{lstlisting}
Answer with only A or B. Do not add any explanation or additional text.

Answer:
\end{lstlisting}
This instruction was originally introduced to simplify analysis of generated responses. However, all reported evaluations use logits-based analysis over the answer tokens (``A'' and ``B''), thereby avoiding confounding effects arising from decoding variability.

\subsection{Example Prompts}
\label{sec:appendix_prompt_examples}
Example prompt illustrating the healthcare domain framing and a temporal recency conflict, where the textual evidence is more recent and aligned with the ground-truth label:
\begin{scriptsize}
\begin{lstlisting}
The patient has a deterioration risk between 0 and 1, where higher values indicate greater risk of clinical deterioration and lower values indicate lower risk.

You are given two timestamped sources describing risk over past time periods: a text summary of risk observations and a time series of numerical risk observations. The two sources may refer to different time windows, and one source may be earlier or more recent than the other. The task is to predict whether the patient's deterioration risk at the target time point will be high or low.

Time series:
14:02: 0.74
14:03: 0.78
14:04: 0.86
14:05: 0.87
14:06: 0.88
14:07: 0.97
14:08: 0.98
14:09: 0.96

Text summary:
"Initially, at 14:26, deterioration risk was at a relatively low level; from there, it increased steadily, and as of 14:33 it was still under control, settling at a relatively low level."

Question: Based on the information above, will the patient's deterioration risk at time 14:34 be high (> 0.5) or low (< 0.5)?

A) high
B) low

Answer with only A or B. Do not add any explanation or additional text.
Answer:
\end{lstlisting}
\end{scriptsize}

Example prompt illustrating the industrial domain framing and a reliability conflict, where the textual evidence is more reliable and aligned with the ground-truth label:
\begin{scriptsize}
\begin{lstlisting}
The machine has a failure risk between 0 and 1, where higher values indicate greater risk of failure, and lower values indicate lower risk.

You are given two sources describing risk over the same time window: a text summary of risk observations and a time series of numerical risk observations. The task is to predict whether the machine's failure risk at the target time point will be high or low.

Time series:
16:50: 0.36
16:51: nan
16:52: nan
16:53: nan
16:54: 0.19
16:55: 0.15
16:56: 0.16
16:57: nan

Text summary:
"Starting at 16:50, failure risk was at an elevated level; from there, it declined gradually, and as of 16:57 it was still concerning, ending at a high level."

Question: Based on the information above, will the machine's failure risk at time 16:58 be high (> 0.5) or low (< 0.5)?

A) high
B) low

Answer with only A or B. Do not add any explanation or additional text.
Answer:
\end{lstlisting}
\end{scriptsize}

Example prompt illustrating the industrial domain framing and a reliability conflict, where the numerical evidence is more reliable and aligned with the ground-truth label:
\begin{scriptsize}
\begin{lstlisting}
The machine has a failure risk between 0 and 1, where higher values indicate greater risk of failure, and lower values indicate lower risk.

You are given two sources describing risk over the same time window: a text summary of risk observations and a time series of numerical risk observations. The task is to predict whether the machine's failure risk at the target time point will be high or low.

Text summary:
"At 14:29, failure risk was at a level just below the critical level; from there, it rose at a moderate pace, and at 14:36 it rose into an elevated state, finishing at a notably high level. Note: this report was produced using partially corrupted data from the observation window."

Time series:
14:29: 0.23
14:30: 0.22
14:31: 0.14
14:32: 0.15
14:33: 0.15
14:34: 0.16
14:35: 0.14
14:36: 0.16

Question: Based on the information above, will the machine's failure risk at time 14:37 be high (> 0.5) or low (< 0.5)?

A) high
B) low

Answer with only A or B. Do not add any explanation or additional text.
Answer:
\end{lstlisting}
\end{scriptsize}

Example prompt illustrating the finance domain framing and a tool forecast conflict, where the contextual evidence (text) is aligned with the ground-truth label:
\begin{scriptsize}
\begin{lstlisting}

The asset has a financial distress risk between 0 and 1, where higher values indicate greater risk of financial distress and lower values indicate lower risk.

You are given a text summary of risk observations over a past time period. An external forecasting tool has analyzed these same observations and produced a predicted risk value at a future time point. The task is to predict whether the asset's financial distress risk at the target time point will be high or low.

Text summary:
"Starting at 14:06, financial distress risk was at a notably high level; after that, it rose gradually, and at 14:13 it was still concerning, settling at an elevated level."

External forecasting tool risk prediction at time 14:14: 0.27

Question: Based on the information above, will the asset's financial distress risk at time 14:14 be high (> 0.5) or low (< 0.5)?

A) high
B) low

Answer with only A or B. Do not add any explanation or additional text.
Answer:

\end{lstlisting}
\end{scriptsize}

\section{Experimental Details}
\label{sec:appendix_exp}
The default configuration of hyperparameters used in main arbitration experiments is reported in Table~\ref{tab:default_config}. All experiments were conducted using the HuggingFace Transformers library and PyTorch. Models were evaluated in inference-only mode using greedy decoding (\texttt{do\_sample=False}) with a maximum generation length of five tokens. Final predictions were derived from the logits of the next-token distribution over the answer tokens ``A'' and ``B'', rather than from generated text. To ensure consistent token indexing across models, we verified that both answer options corresponded to single tokenizer tokens (including leading whitespace). Inference was performed with left-padded inputs and batch size 8. Models were executed in FP16 precision on GPU. Generated responses were stored only for qualitative inspection and were not used for evaluation. All experiments were run on a single NVIDIA RTX PRO 5000 Blackwell GPU (48GB VRAM), with a total computational cost of approximately 40 GPU hours.

\begin{table}[t]
    \centering
    \small
    \begin{tabular}{ll}
        \toprule
        \textbf{Hyperparameter} & \textbf{Default Value} \\
        \midrule
        Answer choices order & ``A'' first \\
        Answer choices-labels mapping & ``A''=\textsc{high} \\
        Sequence length & 16 \\
        Forecast horizon & $k=1$ \\
        Frequency & 1 minute \\
        Domain & \texttt{generic} \\
        Noise standard deviation & 0.05 \\
        Margin range & $[0.35, 0.45]$ \\
        \bottomrule
    \end{tabular}\vspace{2.0mm}
    \caption{\textbf{Default hyperparameter configuration used in the main experiments unless otherwise specified.}}
    \label{tab:default_config}
\end{table}

\subsection{Artifact Usage}
We evaluated the following publicly available pretrained models obtained from the Hugging Face Hub (\url{https://huggingface.co/}):
\begin{itemize}
    \item Qwen3 1.7B, 4B, 8B, 14B (Apache License 2.0);
    \item Gemma-2-9B-It (Gemma Terms of Use (Google));
    \item Llama-3-8B-Instruct (Meta Llama 3 Community License Agreement);
    \item Mistral-7B-Instruct-v0.3 (Apache License 2.0).
\end{itemize}

All models were used for inference-only evaluation without modification or redistribution, consistent with the intended use described in their respective model cards and licenses.

\section{Sensitivity Analyses}
\label{sec:appendix_sensitivity}

Results on sensitivity analyses are reported in Table~\ref{tab:sensitivity}. We evaluate robustness to variations in data-generation parameters (time series length ($T$): 8, 16, 32; noise standard deviation ($\sigma$): 0.01, 0.05, 0.1; margin ranges: $(0.15,0.25)$, $(0.25,0.35)$, $(0.35,0.45)$) and prompt-related parameters (domain specialization: generic, healthcare, finance, industrial; and answer choice configuration, i.e., label semantics and answer ordering). The baseline configuration uses the default values in Table~\ref{tab:default_config}. Reported values correspond to the absolute accuracy change relative to this baseline, aggregated across sweep values, seeds, and evidence-order settings (where applicable) and presented as mean $\pm$ standard deviation. Lower values indicate lower sensitivity to the sweep parameter.

Text-only settings remain highly stable across all sweeps, with most models exhibiting near-zero sensitivity. In contrast, numeric-only settings are generally more sensitive, particularly for \qwenone, \llama, and \mistral, with margin range perturbations producing the largest effects. Conflict settings show more heterogeneous behaviour: some models, especially \qwenfour and \gemma, exhibit substantial sensitivity to time series length despite stable unimodal performance. Across all sweeps, robustness also tends to improve with scale within \qwenfamily, with the 14B model remaining consistently stable.

\begin{table*}[t]
\centering
\small
\begin{tabular}{lcccc}
\toprule
Model 
& \shortstack{$|\Delta \mathrm{Acc}|$\\numeric-only} 
& \shortstack{$|\Delta \mathrm{Acc}|$\\text-only} 
& \shortstack{$|\Delta \mathrm{Acc}|$\\conflict (numeric GT)} 
& \shortstack{$|\Delta \mathrm{Acc}|$\\conflict (text GT)} \\
\midrule
\textbf{Domain Sensitivity} \\
\addlinespace[2pt]
\qwenone      & $0.06 \pm 0.01$ & $0.01 \pm 0.01$ & $0.13 \pm 0.07$ & $0.14 \pm 0.06$ \\
\qwenfour     & $0.00 \pm 0.00$ & $0.00 \pm 0.00$ & $0.05 \pm 0.04$ & $0.05 \pm 0.04$ \\
\qweneight    & $0.00 \pm 0.00$ & $0.00 \pm 0.00$ & $0.05 \pm 0.07$ & $0.05 \pm 0.07$ \\
\qwenfourteen & $0.00 \pm 0.00$ & $0.00 \pm 0.00$ & $0.03 \pm 0.03$ & $0.03 \pm 0.03$ \\
\gemmafull    & $0.00 \pm 0.00$ & $0.00 \pm 0.00$ & $0.04 \pm 0.03$ & $0.04 \pm 0.02$ \\
\llamafull    & $0.09 \pm 0.06$ & $0.00 \pm 0.00$ & $0.03 \pm 0.03$ & $0.04 \pm 0.03$ \\
\mistralfull  & $0.18 \pm 0.08$ & $0.00 \pm 0.00$ & $0.07 \pm 0.03$ & $0.06 \pm 0.03$ \\
\midrule

\addlinespace[2pt]
\textbf{Answer Choice Configuration Sensitivity} \\
\addlinespace[2pt]

\qwenone      & $0.11 \pm 0.16$ & $0.02 \pm 0.01$ & $0.06 \pm 0.07$ & $0.06 \pm 0.07$ \\
\qwenfour     & $0.01 \pm 0.01$ & $0.01 \pm 0.02$ & $0.20 \pm 0.10$ & $0.20 \pm 0.10$ \\
\qweneight    & $0.01 \pm 0.02$ & $0.00 \pm 0.00$ & $0.11 \pm 0.15$ & $0.10 \pm 0.14$ \\
\qwenfourteen & $0.00 \pm 0.00$ & $0.00 \pm 0.00$ & $0.05 \pm 0.04$ & $0.06 \pm 0.04$ \\
\gemmafull    & $0.00 \pm 0.00$ & $0.00 \pm 0.00$ & $0.03 \pm 0.02$ & $0.03 \pm 0.02$ \\
\llamafull    & $0.14 \pm 0.08$ & $0.01 \pm 0.01$ & $0.04 \pm 0.04$ & $0.04 \pm 0.04$ \\
\mistralfull  & $0.22 \pm 0.16$ & $0.00 \pm 0.00$ & $0.17 \pm 0.15$ & $0.18 \pm 0.15$ \\
\midrule

\addlinespace[2pt]
\textbf{Margin Range Sensitivity} \\
\addlinespace[2pt]

\qwenone      & $0.12 \pm 0.05$ & $0.03 \pm 0.02$ & $0.12 \pm 0.04$ & $0.12 \pm 0.05$ \\
\qwenfour     & $0.02 \pm 0.01$ & $0.02 \pm 0.02$ & $0.09 \pm 0.05$ & $0.08 \pm 0.04$ \\
\qweneight    & $0.01 \pm 0.01$ & $0.00 \pm 0.00$ & $0.08 \pm 0.03$ & $0.06 \pm 0.03$ \\
\qwenfourteen & $0.02 \pm 0.02$ & $0.01 \pm 0.01$ & $0.05 \pm 0.04$ & $0.04 \pm 0.04$ \\
\gemmafull    & $0.08 \pm 0.05$ & $0.01 \pm 0.01$ & $0.07 \pm 0.04$ & $0.06 \pm 0.03$ \\
\llamafull    & $0.12 \pm 0.04$ & $0.01 \pm 0.01$ & $0.02 \pm 0.02$ & $0.02 \pm 0.02$ \\
\mistralfull  & $0.14 \pm 0.06$ & $0.00 \pm 0.00$ & $0.03 \pm 0.02$ & $0.02 \pm 0.02$ \\
\midrule

\addlinespace[2pt]
\textbf{Noise Standard Deviation Sensitivity} \\
\addlinespace[2pt]

\qwenone      & $0.06 \pm 0.02$ & $0.01 \pm 0.01$ & $0.07 \pm 0.04$ & $0.06 \pm 0.04$ \\
\qwenfour     & $0.01 \pm 0.01$ & $0.00 \pm 0.00$ & $0.07 \pm 0.03$ & $0.06 \pm 0.02$ \\
\qweneight    & $0.00 \pm 0.00$ & $0.00 \pm 0.00$ & $0.06 \pm 0.05$ & $0.05 \pm 0.04$ \\
\qwenfourteen & $0.00 \pm 0.00$ & $0.00 \pm 0.00$ & $0.03 \pm 0.02$ & $0.02 \pm 0.02$ \\
\gemmafull    & $0.02 \pm 0.01$ & $0.00 \pm 0.00$ & $0.04 \pm 0.02$ & $0.04 \pm 0.03$ \\
\llamafull    & $0.10 \pm 0.05$ & $0.00 \pm 0.00$ & $0.06 \pm 0.06$ & $0.05 \pm 0.06$ \\
\mistralfull  & $0.09 \pm 0.07$ & $0.00 \pm 0.00$ & $0.04 \pm 0.03$ & $0.04 \pm 0.03$ \\
\midrule

\addlinespace[2pt]
\textbf{Time Series Length Sensitivity} \\
\addlinespace[2pt]

\qwenone      & $0.03 \pm 0.01$ & $0.01 \pm 0.00$ & $0.07 \pm 0.05$ & $0.08 \pm 0.06$ \\
\qwenfour     & $0.00 \pm 0.00$ & $0.00 \pm 0.00$ & $0.13 \pm 0.08$ & $0.12 \pm 0.08$ \\
\qweneight    & $0.00 \pm 0.00$ & $0.00 \pm 0.00$ & $0.06 \pm 0.04$ & $0.06 \pm 0.03$ \\
\qwenfourteen & $0.00 \pm 0.00$ & $0.00 \pm 0.00$ & $0.04 \pm 0.02$ & $0.04 \pm 0.03$ \\
\gemmafull    & $0.00 \pm 0.00$ & $0.00 \pm 0.00$ & $0.16 \pm 0.04$ & $0.16 \pm 0.04$ \\
\llamafull    & $0.07 \pm 0.07$ & $0.00 \pm 0.00$ & $0.06 \pm 0.06$ & $0.06 \pm 0.06$ \\
\mistralfull  & $0.03 \pm 0.03$ & $0.00 \pm 0.00$ & $0.08 \pm 0.11$ & $0.08 \pm 0.12$ \\

\bottomrule
\end{tabular}\vspace{2.0mm}
\caption{\textbf{Absolute accuracy change ($|\Delta \mathrm{Acc}|$) under different sensitivity-analysis settings.} Accuracy deltas are aggregated across sweep values, seeds, and (where applicable) evidence-order settings, and reported as mean $\pm$ standard deviation. Lower values indicate lower sensitivity to the swept parameter. GT: ground truth.}
\label{tab:sensitivity}
\end{table*}

\end{document}